\documentclass[conference,a4paper]{IEEEtran}
\IEEEoverridecommandlockouts

\usepackage[hidelinks]{hyperref}
\usepackage[cmex10]{amsmath}
\usepackage{amssymb,amsfonts}
\usepackage{dblfloatfix}

\usepackage[ruled,vlined]{algorithm2e}
\usepackage{graphicx}
\graphicspath{{sec/img/}{Figures/PNG/}}

\usepackage{booktabs}
\usepackage{siunitx}
\usepackage[numbers,compress]{natbib}
\usepackage{texnames}
\usepackage{bm,bbm}

\usepackage{subcaption}
\usepackage[dvipsnames]{xcolor}
\usepackage{orcidlink}

\newcommand{\BenchmarkName}{ForTy}
\newcommand{\MethodName}{MTSViT}

\begin{document}

\title{\uppercase{Not every tree is a forest: benchmarking forest types from satellite remote sensing}}

\author{	\IEEEauthorblockN{Yuchang Jiang\orcidlink{0000-0001-7293-1304}${}^{*}$\thanks{${}^{*}$This work was done while being an intern at Google DeepMind.}}
	\IEEEauthorblockA{\textit{University of Zurich}\\
		yuchang.jiang@uzh.ch}
	\and
	\IEEEauthorblockN{Maxim Neumann\orcidlink{0000-0002-1967-318X}}
	\IEEEauthorblockA{\textit{Google DeepMind}\\
		maximneumann@google.com}
}

\maketitle
\begin{abstract}
	Developing accurate and reliable models for forest types mapping is critical to support efforts for halting deforestation and for biodiversity conservation (such as European Union Deforestation Regulation (EUDR)). This work introduces \uppercase{\BenchmarkName}, a benchmark for global-scale \uppercase{for}est \uppercase{ty}pes mapping using multi-temporal satellite data\footnote{\BenchmarkName\ (v1) is released under CC-BY-SA 4.0 license. Access instructions at \href{https://github.com/google-deepmind/forest_typology}{github.com/google-deepmind/forest\_typology}.}. The benchmark comprises 200,000 time series of image patches, each consisting of Sentinel-2, Sentinel-1, climate, and elevation data. Each time series captures variations at monthly or seasonal cadence. Per-pixel annotations, including forest types and other land use classes, support image segmentation tasks. Unlike most existing land use products that often categorize all forest areas into a single class, our benchmark differentiates between three forest types classes: natural forest, planted forest, and tree crops. By leveraging multiple public data sources, we achieve global coverage with this benchmark. We evaluate the forest types dataset using several baseline models, including convolution neural networks and transformer-based models. Additionally, we propose a novel transformer-based model specifically designed to handle multi-modal, multi-temporal satellite data for forest types mapping. Our experimental results demonstrate that the proposed model surpasses the baseline models in performance.
\end{abstract}


\section{Introduction}

Forests play a crucial role in mitigating climate change while providing social, economic, and cultural benefits. It is especially important to differentiate between natural forests, planted forests, and tree crops to create actionable data supporting policy-making, conservation, and sustainable forest management \citep{richter2024SDPTv2}.
However, AI and computer vision research mostly focuses on forest recognition, differentiating only between forested and non-forested areas \citep{turubanova2018ongoing_primary2001, sabatini2021european_euPrimary, zanaga2021esa_worldcover2020, brown2022_dynamicWorld}. For instance, forest plantations for rubber or timber are still classified as forest cover\cite{zalles2024forestrethink}. While these maps provide a basic overview, they are inadequate for addressing deforestation risks and monitoring degradation, particularly under regulations like the European Union Deforestation Regulation (EUDR).

%
%

Satellite imagery and remote sensing, combined with machine learning, have become vital tools for forest monitoring. Foundation models trained on large-scale datasets are gaining traction in remote sensing but primarily capture general satellite image representations \citep{jakubik2023Prithvi, fuller2024croma}. A large-scale benchmark tailored for forest types mapping could advance this field and enable the development of forest-specific foundation models.
While existing forest-related machine learning benchmarks provide valuable resources, they have notable limitations. For instance, BigEarthNet \citep{sumbul2019bigearthnet} is a large dataset designed for deep learning but focuses on general land-cover classification rather than forest type mapping and provides only per-image labels. Regionally constrained datasets, such as TreeSatAI \citep{ahlswede2022treesatai} and BioMassters \citep{nascetti2024biomassters}, are tailored to specific forest-related tasks but are not designed for forest type mapping. A more closely related dataset is Planted \citep{pazos2024planted}, which utilizes the Spatial Database of Planted Trees (SDPT) \citep{richter2024SDPTv2} (a collection of data sources, including among others \citep{, Lesiv2022:gfm, roy2015:india, pan2011:noam-forests, furumo2017:oilpalm, petersen2016:tree-plantations}) to identify planted forests and tree crops. However, it also provides only per-image labels and lacks negative samples for non-planted forests. These limitations underscore the need for a comprehensive benchmark that enables detailed forest types mapping with per-pixel labels on a global scale.

We introduce \BenchmarkName, a new global-scale, multi-modal multi-temporal benchmark  dataset designed to advance research in forest types mapping. \BenchmarkName\ features 200,000 globally distributed locations annotated with per-pixel labels and includes detailed forest classes: natural forests, planted forests, and tree crops, alongside non-forest categories. We propose a new transformer-based model (\MethodName) for forest type mapping and provide initial baselines results for the benchmark.

\section{The \BenchmarkName\ dataset}


\begin{figure}[b]
	\centering
	\vspace{-1em}
    \includegraphics[width=0.8\linewidth]{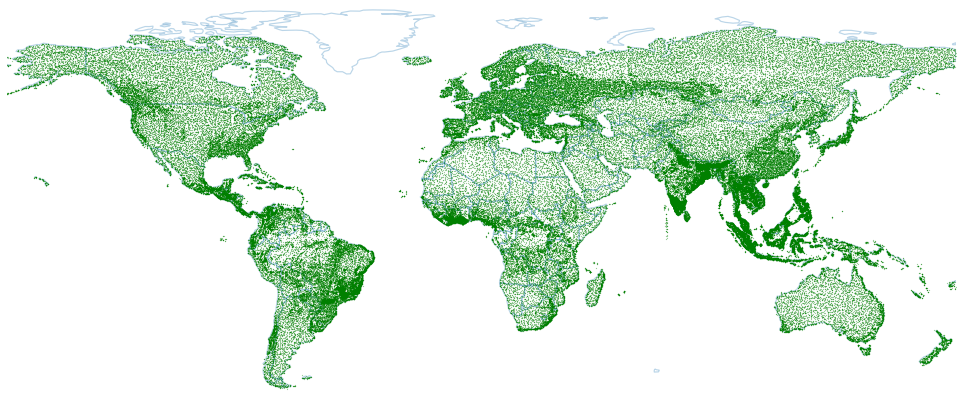}
    \caption{\BenchmarkName\ dataset stratified random sample locations.}
    \label{fig:coverage}
\end{figure}

To enable global wall-to-wall mapping, our reference data contain 8 classes: three forest classes (natural forest, planted forest, tree crops), other vegetation, water, ice, bare ground, and built areas. 
For \textit{natural} forest class, we integrate data from diverse sources to achieve a global coverage: primary humid tropical forests \citep{turubanova2018ongoing_primary2001}, primary forests in Europe \citep{sabatini2021european_euPrimary}, Canada \citep{maltman2023estimating_caPrimary}, and USA \citep{dellasala2022mature_usaPrimary}, natural forests from tropical moist forests (TMF) \citep{vancutsem2021long_TMF}, and natural lands from Science Based Targets Network (SBTN) \citep{Mazur_Sims_Goldman_Schneider_Pirri_Beatty_Stolle_Stevenson_SBTN}.
For \textit{planted} forest class, we merge planted forests from the planted areas from TMF \citep{vancutsem2021long_TMF} and SDPT \citep{richter2024SDPTv2}.
The \textit{tree crop} class integrates data from SDPT \citep{richter2024SDPTv2}, CORINE Land Cover \citep{cover2018copernicus_CORINE}, and tree crop species in cropland from national agricultural statistics service (USDA) \citep{nass2016usda} with several regional tree crop datasets covering a diverse range of tree crops, including palm, cocoa, rubber, coffee, coconut, cashew, and other orchards  \citep{kalischek2022_cocoaETH,  danylo2021_palmDanylo, vollrath2020angular_palmVollrath, descals2024global_palmDescals, fricker2022palmUcayali, souza2020mapbiomas, descals2023high_coconutDescals, sheil2024rubber, sourcecooperative2024_cashewBenin}.
The \textit{other vegetation} class includes areas identified as shrubland, grassland, and cropland in WorldCover \citep{zanaga2021esa_worldcover2020} and areas labeled as vegetation in SBTN \citep{Mazur_Sims_Goldman_Schneider_Pirri_Beatty_Stolle_Stevenson_SBTN}. Only pixels confirmed by both data sources are assigned to the other vegetation class.
For non-vegetation classes (\textit{water}, \textit{ice}, \textit{bare ground}, \textit{built areas}), we use labels from World Cover \citep{zanaga2021esa_worldcover2020}.

When integrating the three forest layers, we address disagreements between forest types by assigning a forest label only when there is consensus among the layers. In cases of disagreement, the pixels are labeled as \textit{unknown}. To incorporate non-forest classes, we use a tree height map \citep{potapov2022global_treemask}, adding non-forest labels only to areas where the tree height is less than 5~m. For regions affected by deforestation with high confidence of tree regrowth based on identified drivers \citep{sims2024drivers}, we assign planted forest label if no other valid label is applied.
We acknowledge the presence of noise in these data sources. Our goal is to develop methods that can effectively handle such noise, reflecting the complexities of real-world scenarios.

\subsection{Satellite Data}
\begin{figure}
	\centering
    \includegraphics[width=\linewidth]{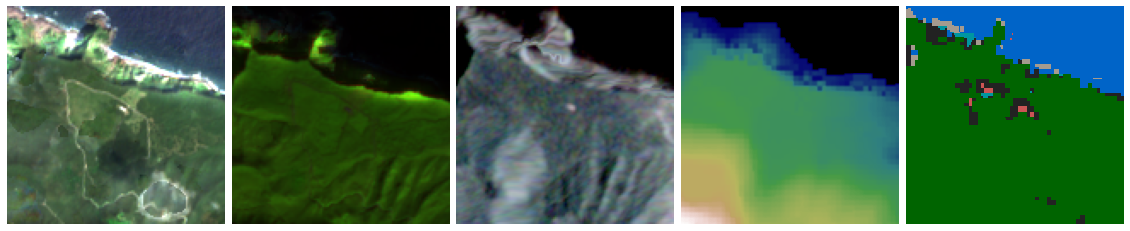}\\
    \includegraphics[width=\linewidth]{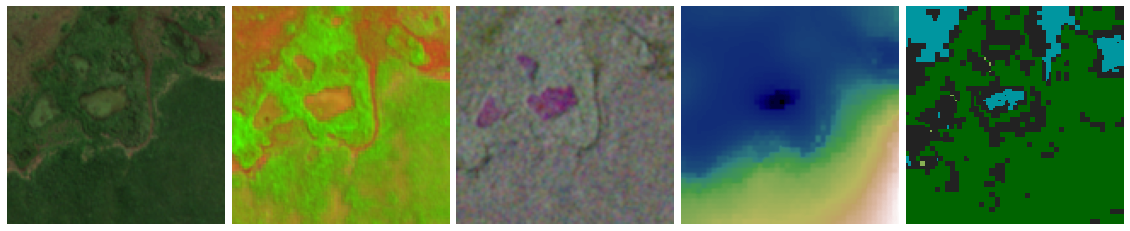}\\
    \includegraphics[width=\linewidth]{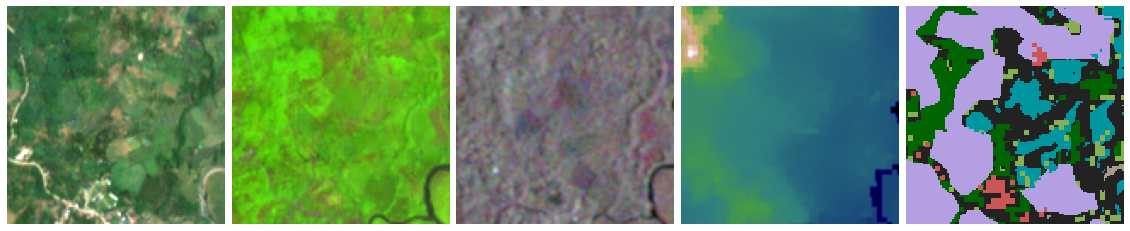}\\
    \includegraphics[width=\linewidth]{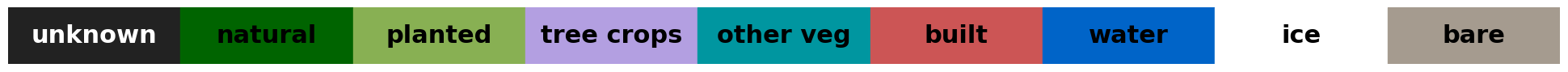}\\
    \caption{Examples from left to right: (1) Sentinel-2 RGB bands, (2) Sentinel-2 SWIR-NIR-Red bands, (3) multi-temporal Sentinel-1 VV winter-spring-summer composition, (4) elevation, (5) labels. Bottom row: color bar for labels.\vspace{-0.9em}}
    \label{fig:examples}
\end{figure}

This benchmark dataset includes mutli-modal satellite inputs: Sentinel-2, Sentinel-1, climate, and elevation data (examples are shown in \autoref{fig:examples}).
\begin{itemize}
    \item \textbf{Sentinel-2} provides multispectral optical satellite imagery. For this benchmark, we extract seasonal and monthly mosaics from Sentinel-2 imagery for up to three years (2018-2020). We use 10 spectral bands with nominal resolutions of 10~m and 20~m, which are particularly useful for land cover mapping.
    \item \textbf{Sentinel-1} is a Synthetic Aperture Radar (SAR) satellite with 10~m resolution. Similar to Sentinel-2, we create mosaics for three years and use VV and VH polarizations from both ascending and descending orbits.
    \item \textbf{Climate} \citep{abatzoglou2018terraclimate} is a monthly climate dataset with a spatial resolution of approximately 4~km. This dataset provides key climate and water balance variables globally.
    \item \textbf{Elevation} is sourced from FABDEM \citep{hawker2022FABDEM}, based on Copernicus GLO 30 DEM with forests and buildings removed, providing elevation, slope, and aspect.
\end{itemize}

\subsection{Dataset Construction}

We use random stratified sampling to get locations distributed across the globe (see Figure~\ref{fig:coverage}), prioritizing forest type classes to ensure sufficient numbers of plots and pixels. The final dataset contains about 200,000 plots of 1280 $\times$ 1280 meter sizes. For each location, we extract satellite imagery and labels centered on the location.
To ensure geographically distinct datasets, we divide the world into $100\times 100\ km^2$ blocks and randomly assign them to the training, validation, and test sets in an 8:1:1 ratio. This approach guarantees that the training, validation, and test sets are geographically separated, reducing spatial autocorrelation and ensuring robust model evaluation.

\subsection{Dataset Analysis}
\begin{figure}[hbt]
	\centering
    \begin{subfigure}[b]{0.45\textwidth}
        \centering
        \includegraphics[width=\textwidth]{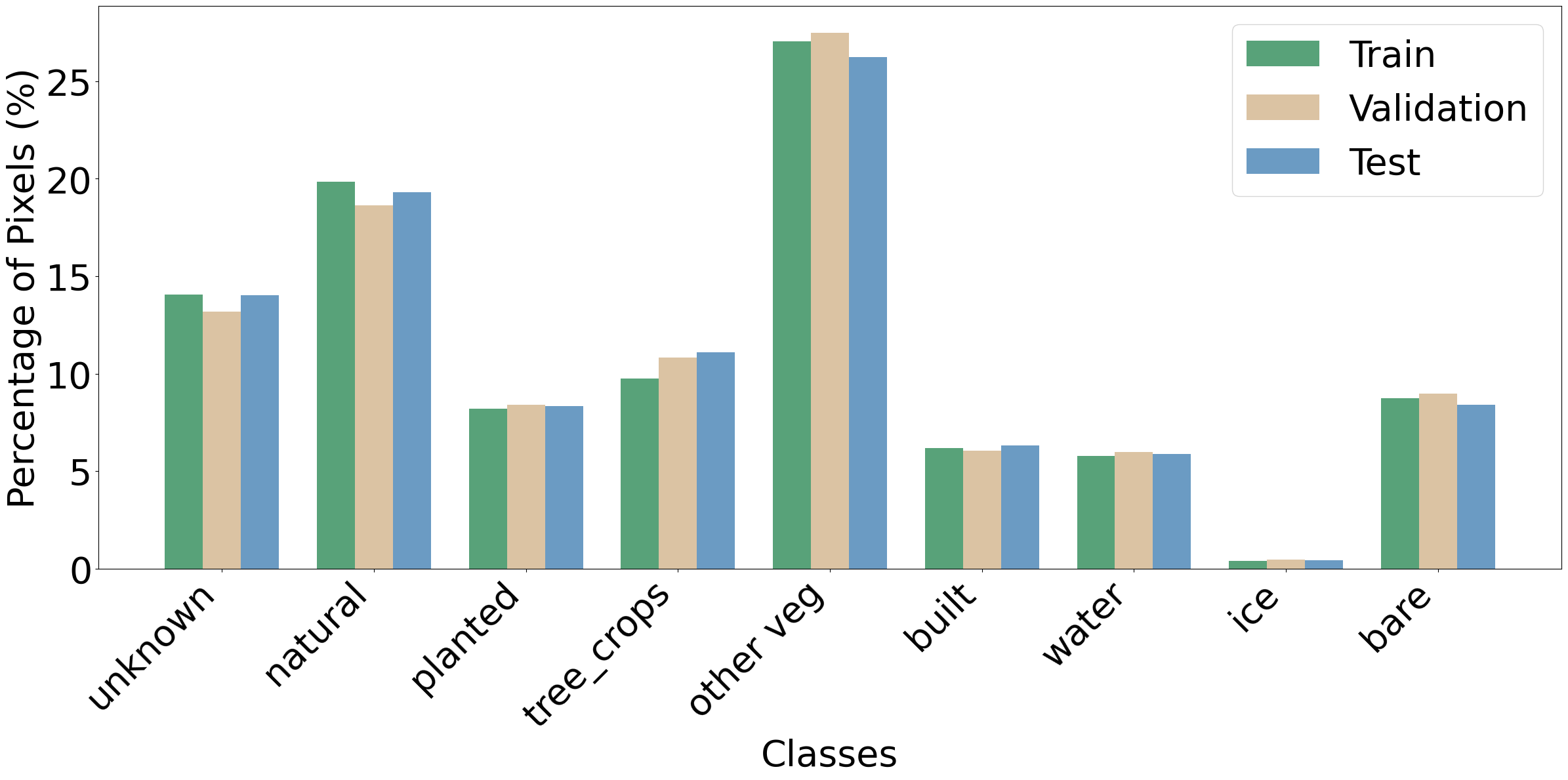}
        \caption{Distribution of pixels across classes.}
        \label{fig:number_of_pixels}
    \end{subfigure}
    \hfill
    
    \begin{subfigure}[b]{0.45\textwidth}
        \centering
        \includegraphics[width=\textwidth]{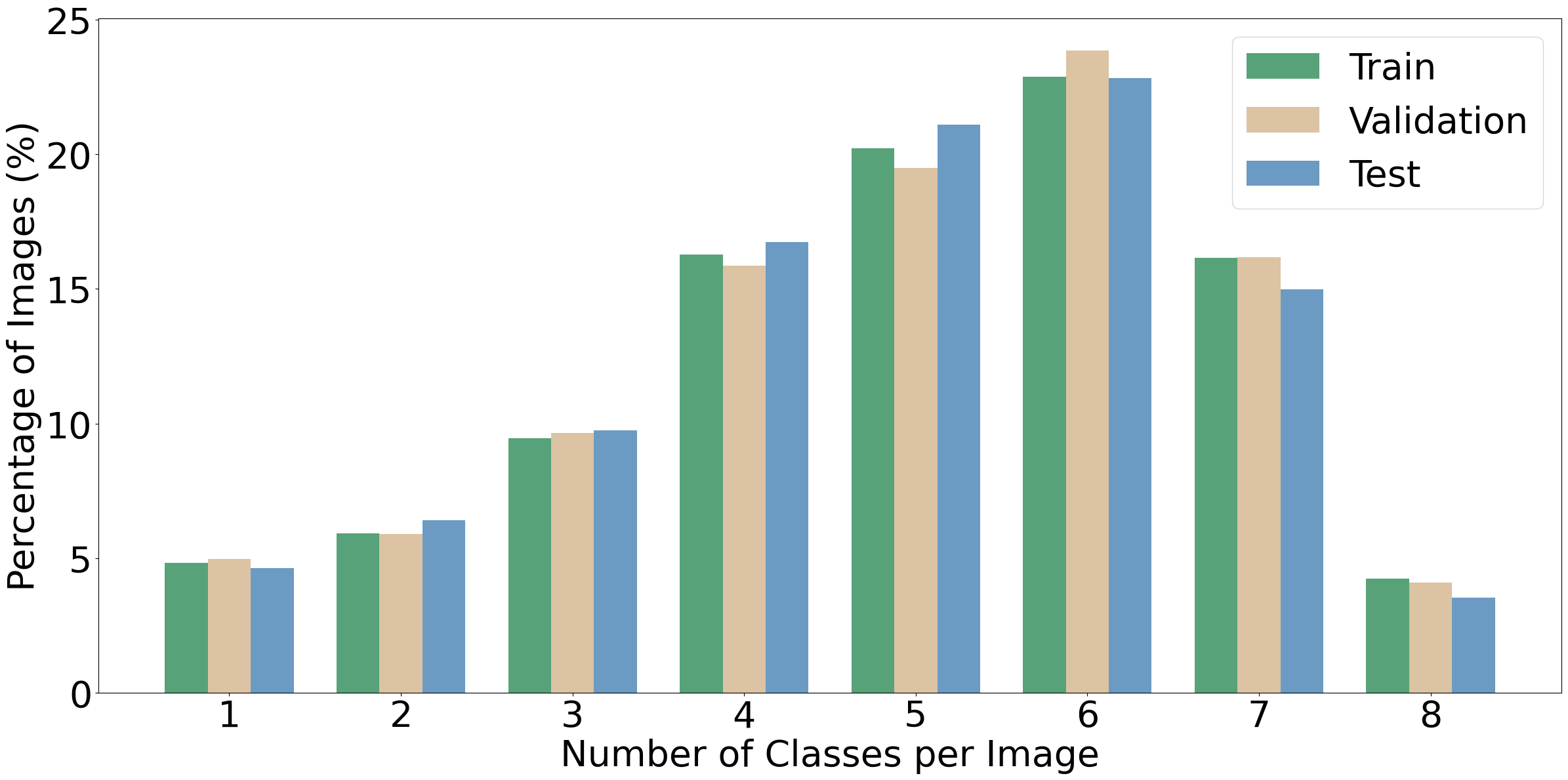}
        \caption{Distribution of number of classes per image.}
        \label{fig:number_of_classes}
    \end{subfigure}
    
    \caption{Overview of class distributions: (a) shows the distribution of pixels across classes, while (b) shows the distribution of the number of classes per image.\vspace{-0.9em}}
    \label{fig:class_distributions}
\end{figure}

For the constructed benchmark, we compute the dominant class for each image sample. Approximately 13\% of the images are dominated by natural forest, 10\% by planted forests, 7\% by tree crops, and 17\% by other vegetation across the training, validation, and test sets. To further analyze the dataset, we investigate the class distribution based on the number of pixels and the number of distinct classes in each image, as illustrated in Figure~\ref{fig:class_distributions}.
%
%
In the pixel distribution across classes, we observe that planted forests and tree crops occupy fewer pixels compared to natural forests and other vegetation. While we use stratified sampling to prioritize forest type classes and achieve a more balanced distribution at the plot level, a perfectly balanced distribution at the pixel level is neither expected nor feasible. This imbalance reflects the diversity of the data and mirrors real-world conditions.
In the distribution of the number of classes per image, we find that most image samples contain at least four distinct classes. This indicates that by integrating diverse data sources, our benchmark provides rich segmentation labels, capturing the complexity and diversity of real-world land-cover scenarios.

\section{The \MethodName\ Model}

Attention mechanisms and transformers have become widely adopted in remote sensing applications \citep{garnot2021UTAE, tarasiou2023TSViT}. In this work, we benchmark against convolutional neural network (CNN) models, CNNs enhanced with attention mechanisms, and transformer architectures specifically designed for temporal satellite inputs. Building on Temporal-Spatial Vision Transformer (TSViT) \citep{tarasiou2023TSViT}, a ViT-based model tailored for time-series satellite data, we propose Multi-modal Temporal Spatial Vision Transformer (\textbf{\MethodName}) to handle multi-modal, multi-temporal inputs for per-pixel forest type mapping.

\begin{figure}[t]
	\centering
	\includegraphics[width=\linewidth]{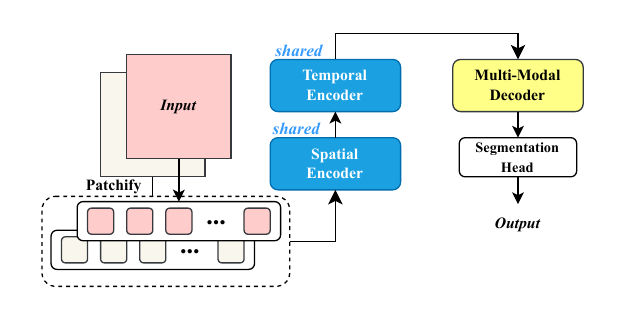}
	\caption{An overview of the proposed model. The example demonstrates the use of two input modalities. The encoders are shared across modalities, while the decoder facilitates multi-modal interaction before the final segmentation head.\vspace{-0em}}\label{fig:model_overview}
\end{figure}

Figure~\ref{fig:model_overview} provides an overview of the model architecture. \MethodName\ consists of a spatial encoder, a temporal encoder, a multi-modal decoder, and a segmentation head to produce the per-pixel segmentation results. Both the encoders and the decoder are lightweight, consisting of two transformer layers, which effectively capture spatial, temporal, and multi-modal interactions.

Following the standard transformer architecture for 3D patches, we patchify and project each satellite modality with the original shape $[B, T_m, H_m, W_m, C_m]$ into an embedding with the shape $[B, t_m, h_m, w_m, d]$. Here, $B$ denotes the batch size. $T_m$, $H_m$, $W_m$, and $C_m$ represent the original temporal, spatial, and channel dimensions of modality $m$, and $t_m$, $h_m$, $w_m$ are the patchified temporal and spatial dimensions of modality $m$. The embedding dimension, $d$, is set to 192.
In both the spatial and temporal encoders, all modalities share the same encoders; thus, we omit the modality notation $m$ in the following description. The tensors are reshaped into $[Bt, hw, d]$ to focus the spatial encoder exclusively on processing tokens from the spatial dimension. For tensors from different modalities with varying $hw$, we pad them to the longest spatial dimension and mask irrelevant tokens in the attention layers. This enables the spatial encoder to handle modalities with differing spatial resolutions. Non-spatial satellite inputs, such as climate data, bypass the spatial encoder.
After processing by the spatial encoder, the tensors are rearranged into $[Bhw, t, d]$, allowing the temporal encoder to focus solely on the temporal dimension. Additionally, we concatenate the sequence with $K$ class queries ($[Bhw, t+K, d]$), following the approach introduced in \cite{tarasiou2023TSViT}. We feed the resulted tensors into a standard transformer encoder. To eliminate the temporal dimension after the temporal encoder, we retain only the $K$ embeddings corresponding to the class queries ($[Bhw, K, d]$) and reshape them into $[BK, hw, d]$ for further processing.

In both the spatial and temporal encoders, we independently process image embeddings from each satellite modality. The multi-modality decoder is specifically designed to enable interaction among different modalities. This is achieved by using the embedding from one satellite modality as queries ($[BK, hw, d]$) and the embeddings from the other modalities as keys and values. For instance, climate and elevation features can be queried based on information from Sentinel-2 optical imagery. This approach leverages the standard transformer decoder to extract complementary features from other modalities guided by the context of a single modality. 
Finally, a Multilayer Perceptron (MLP) is applied to project and reshape the decoder’s output into predictions, preserving the original spatial dimensions of the satellite modality used for querying.

\section{Experimental Results}

\subsection{Experimental Setup}

We evaluate our benchmark using three baseline models: \textbf{UNet3D}, a CNN-based model \citep{rustowicz2019:unet3d}, \textbf{UTAE}, a CNN model with temporal attentions \citep{garnot2021UTAE}, and \textbf{TSViT}, a transformer model specifically designed to process temporal satellite inputs \citep{tarasiou2023TSViT}.

For all baseline models and our proposed approach, we use the same implementation setup to ensure a fair comparison. Input channels are normalized using training set statistics. We train all models using the cross-entropy loss function. All implementations are developed in JAX \citep{jax2018github}. 
We train the models using a batch size of 64 for 40 epochs, with a learning rate of 0.001. We use the Adam optimizer with $\beta_1 = 0.9$, $\beta_2 = 0.9999$, and $\eta = 10^{-8}$.

For the final evaluations, we train all models on the combined training and validation sets and evaluate them on the test set to report final performance metrics. To ensure robustness, we conduct repeated analyses with three different random seeds and report the average metrics. Per trained model, the performance is evaluated using the overall F1 score (macro-averaged across all classes) and the F1 score for forest classes (averaged over the three forest classes). Additionally, we report F1 scores for each forest type class.

\subsection{Main results}

We report the performance of baselines and our proposed \MethodName\ using seasonal Sentinel-2, climate, and elevation from 2020 as inputs, including standard deviations, in Table~\ref{tab:main_results}. Our model achieves the best performance across all metrics. The second-best performer is TSViT, which serves as the foundation for our proposed model. The key difference lies in the addition of a multi-modal decoder in \MethodName\ before the final segmentation head. This enhancement leads to a 1.8\% increase in forest F1 and 1\% to 2.7\% improvements for individual forest classes. CNN-based models, UNet3D and UTAE, show poorer performance, particularly for the planted forest and tree crop classes, which are more challenging due to their imbalanced distribution and smaller spatial coverage.

\begin{table}[hbt]
    \centering
    \caption{Results (F1 score and standard deviation) on the {\t test} set using Sentinel-2, climate, and elevation data over 1 year: \textit{Overall} F1, for the 3 \textit{Forests} classes, and for the individual forest types (natural (\textit{N}), planted (\textit{P}), and tree crops (\textit{TC})).
    }\label{tab:main_results}
    \begin{tabular}{lccccc}
    \toprule
           & \textbf{Overall} & \textbf{Forests} & \textbf{N} & \textbf{P} & \textbf{TC} \\ \midrule
    \textbf{UNet3D}   & 32.4 (1.3)    &24.2 (5.3)           & 56.2 (8.3)          &          7.5 (8.8) &    8.8 (12.1)      \\
    \textbf{UTAE}        &  49.4 (0.8)   & 37.7 (5.0)           &     71.4 (0.0)      &    13.8 (5.4)       &     27.8 (9.6)     \\
    \textbf{TSViT}       &   79.6 (0.1)  &    73.1 (0.1)        &     81.8 (0.0)      &    60.2 (0.3)       &   77.5 (0.2)       \\
    \textbf{Ours}     &  \textbf{81.1} (0.1)   &     \textbf{74.9} (0.1)       &      \textbf{82.8} (0.0)     &    \textbf{62.9} (0.3)       &  \textbf{78.9} (0.2)        \\ \bottomrule
    \end{tabular}
\end{table}

\subsection{Multi-modality ablation}

\begin{figure}[hbt]
	\centering
	\includegraphics[width=.8\linewidth]{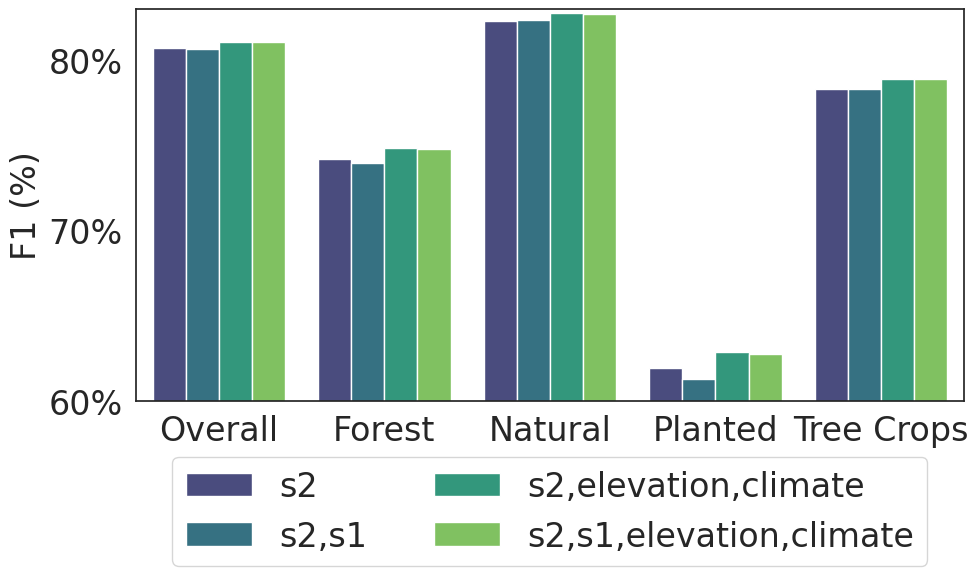}
	\caption{Performance metrics across different satellite inputs combinations.}\label{fig:multimodal}
\end{figure}

We evaluate the impact of different modality combinations on our proposed model using seasonal inputs from the year 2020. Figure~\ref{fig:multimodal} presents the performance across various input combinations. While the overall F1 score remains relatively consistent across inputs, we observe larger differences in the planted and tree crop classes. Incorporating elevation and climate data improves performance across all metrics, whereas the addition of Sentinel-1 shows minimal impact.

\subsection{Multi-temporal ablation}

\begin{figure}[hbt]
	\centering
	\includegraphics[width=.8\linewidth]{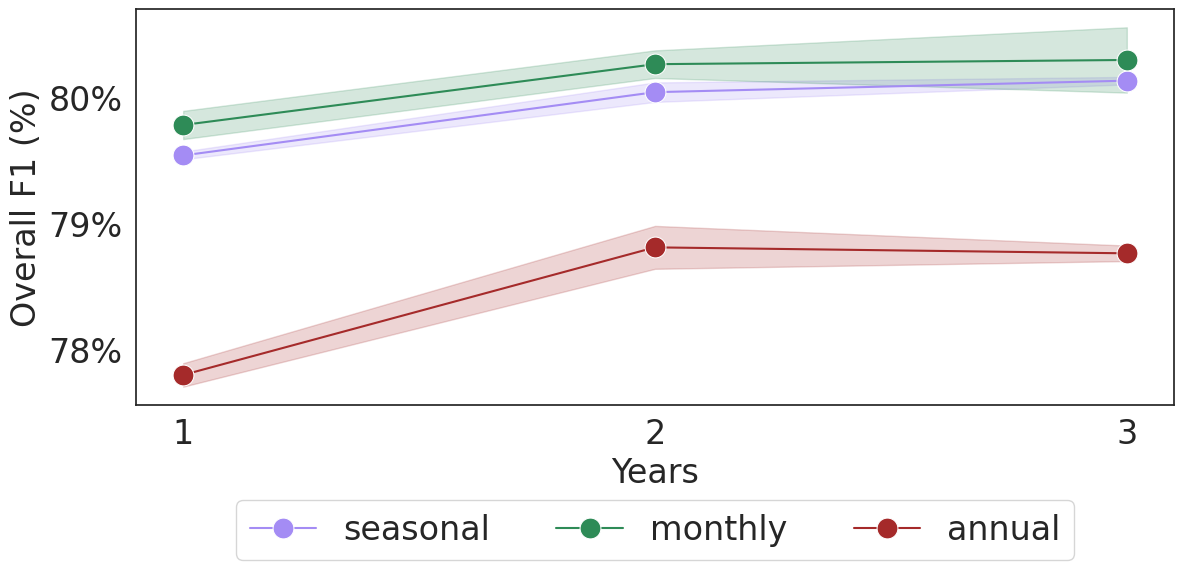}
	\caption{Performance of overall F1 scores across different cadences (seasonal, monthly, and annual) over three years. The y-axis represents the F1 score as a percentage, while the x-axis indicates the number of years used.}\label{fig:temporal}
\end{figure}

Our dataset includes monthly and seasonal satellite inputs spanning three years. We investigate how temporal resolution and the amount of data (in years) impact the results. Figure~\ref{fig:temporal} illustrates the performance of our model using Sentinel-2 data with varying temporal resolutions and years of input. The results clearly show that monthly and seasonal inputs significantly outperform annual inputs. While using more years of data further improves performance, the effect is less pronounced compared to the impact of temporal resolution.

\section{Conclusion}

We present \BenchmarkName, a comprehensive benchmark dataset designed to advance research in forest type mapping. Our dataset goes beyond traditional single-class forest representations by providing detailed forest types, including natural forests, planted forests, and tree crops, along with other land-cover classes. With its global coverage, multi-temporal multi-modal inputs, and per-pixel annotations, \BenchmarkName\ serves as a robust resource for training and evaluating machine learning models.
We also propose a novel model architecture tailored for multi-temporal and multi-modal inputs, effectively handling the complexities of forest type classification. Through evaluations and ablation studies, we compare various model architectures, including CNN and transformer models.
%
In future versions, we plan to incorporate additional satellite modalities, evaluate a broader range of geospatial models, and continuously improve data quality. We hope this dataset encourages researchers to explore challenges in forest type mapping and identification, whether through novel model designs, enhanced temporal and modality representations, or improvements in label construction.



\small
\bibliographystyle{IEEEtranN}
\bibliography{refs}

\end{document}